\documentclass{article}

\usepackage{arxiv}

\usepackage{makecell, tabularx}
\usepackage[export]{adjustbox}

\usepackage{algorithm}
\usepackage{tabularx}
\usepackage{rotating}
\usepackage[algo2e]{algorithm2e} 
\usepackage{algpseudocode}
\usepackage{url}
\usepackage[table,xcdraw]{xcolor}
\usepackage[colorlinks,linkcolor=blue,anchorcolor=blue,citecolor=green,urlcolor=black]{hyperref}
\usepackage[numbers]{natbib}
\usepackage{amsmath,amssymb}
\usepackage{bbm}
\usepackage{xcolor, soul}
\usepackage[utf8]{inputenc}
\usepackage[english]{babel}
\usepackage{graphicx}
\usepackage{booktabs}
\usepackage{multirow}
\usepackage{subcaption}
\usepackage{algpseudocode}
\usepackage{algorithm}
\usepackage{tabularray}
\usepackage[font=footnotesize]{caption}  
\usepackage{bbding}

\newcolumntype{C}[1]{>{\centering\arraybackslash}p{#1}}
\renewcommand{\shorttitle}{SAT-Cap}

\usepackage{authblk}

\author[1,2]{Yuduo Wang}
\author[1,3]{Weikang Yu}
\author[1,4]{Pedram Ghamisi}
\affil[1]{Helmholtz-Zentrum Dresden-Rossendorf (HZDR), 09599 Freiberg, Germany} 
\affil[2]{Humboldt-Universität zu Berlin, 10099 Berlin, Germany}
\affil[3]{Technical University of Munich, 80333 Munich, Germany}
\affil[4]{Lancaster University, LA1 4YR Lancaster, U.K.}

\date{}    
\title{Change Captioning in Remote Sensing: Evolution to SAT-Cap – A Single-Stage Transformer Approach}

\begin{document}
\maketitle
\begin{abstract}
Change captioning has become essential for accurately describing changes in multi-temporal remote sensing data, providing an intuitive way to monitor Earth's dynamics through natural language. However, existing change captioning methods face two key challenges: high computational demands due to multistage fusion strategy, and insufficient detail in object descriptions due to limited semantic extraction from individual images. To solve these challenges, we propose SAT-Cap based on the transformers model with a single-stage feature fusion for remote sensing change captioning. In particular, SAT-Cap integrates a Spatial-Channel Attention Encoder, a Difference-Guided Fusion module, and a Caption Decoder. Compared to typical models that require multi-stage fusion in transformer encoder and fusion module, SAT-Cap uses only a simple cosine similarity-based fusion module for information integration, reducing the complexity of the model architecture. By jointly modeling spatial and channel information in Spatial-Channel Attention Encoder, our approach significantly enhances the model's ability to extract semantic information from objects in multi-temporal remote sensing images. Extensive experiments validate the effectiveness of SAT-Cap, achieving CIDEr scores of 140.23\% on the LEVIR-CC dataset and 97.74\% on the DUBAI-CC dataset, surpassing current state-of-the-art methods. The code and pre-trained models will be available at \href{https://github.com/AI4RS/SAT-Cap}{https://github.com/AI4RS/SAT-Cap}.
\end{abstract}


\section{Introduction}

With the rapid development of geoscience and remote sensing techniques over the past decades, the collection of Earth observation data has become increasingly accessible. In this era of big data, cutting-edge AI techniques have been employed to analyze imaging data, offering insights into land surface dynamics. Simultaneously, recent advances in natural language processing (NLP) have garnered significant interest for their potential to interpret Earth observation data using text descriptions \cite{li2024vision, kuckreja2024geochat, zhang2024rs5m, liu2024remote, zhang2024earthgpt}. When these advancements converge, integrating image and text modalities not only facilitates AI algorithms by providing a more user-friendly interface for domain experts but also enables richer, multidimensional insights beyond what a single modality can offer. As a result, vision-language models have emerged as a pivotal research focus, advancing the understanding of Earth observation data within the geoscience and remote sensing community.

Vision-language models can enhance the interpretation of Earth observation data in two key ways: through the input and output of AI algorithms in geoscience and remote sensing applications. On one hand, these models can generate textual outputs that are more accessible and comprehensible to a broader audience, including those with basic domain knowledge \cite{zhan2023rsvg, cheng2022nwpu}. On the other hand, textual information provided by users can serve as supplementary input alongside Earth observation data, enabling the models to deliver more precise predictions \cite{liu2024remoteclip, dong2024changeclip}. In geoscience and remote sensing, vision-language models have been applied to tasks such as image captioning \cite{vinyals2015show, zhao2021high, chen2017sca}, text-to-image generation \cite{mansimov2015generating, xu2023txt2img, liu2025text2earth}, visual question answering (VQA) \cite{antol2015vqa, wang2024rsadapter, lobry2020rsvqa, zhang2023spatial}, and image-text retrieval \cite{yan2015deep, yuan2022remote, cheng2021deep}. 

\begin{figure*}
    \centering
    \includegraphics[width=\linewidth]{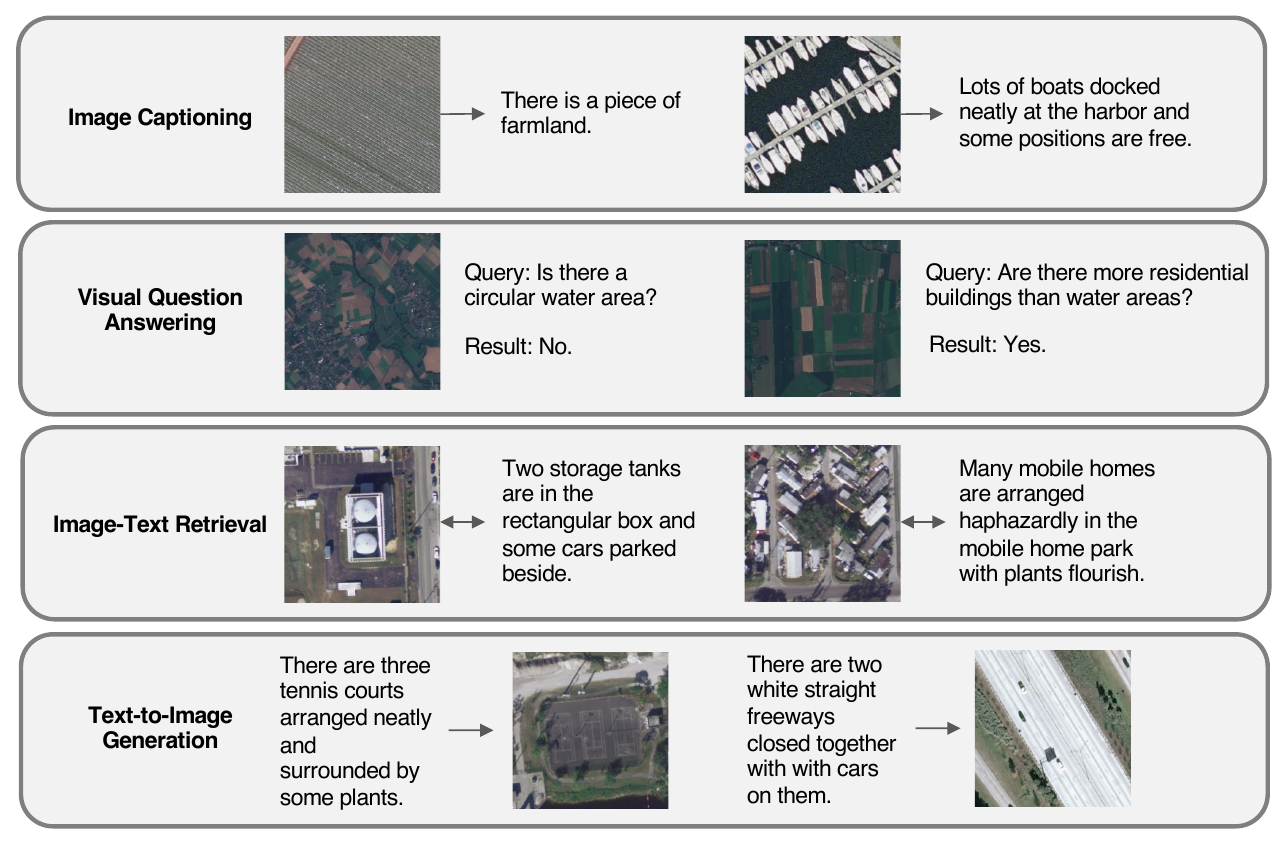}
    \caption{Examples of common remote sensing vision-language downstream tasks: (a) image captioning, (b) visual question answering, (c) image-text retrieval, and (d) text-to-image generation}
    \label{fig:1_1}
\end{figure*}

Figure \ref{fig:1_1} illustrates examples of four distinct remote sensing vision-language downstream tasks. Image captioning generates descriptive text that explains the content of an image, effectively translating visual information into a narrative. Text-to-image generation performs the reverse operation, synthesizing realistic images from textual descriptions using generative models such as GANs \cite{reed2016generative} or diffusion models \cite{croitoru2023diffusion}. Image-text retrieval involves identifying the most relevant images based on a textual query (text-to-image retrieval) or finding the most relevant textual descriptions for a given image (image-to-text retrieval). Visual question answering (VQA) requires a system to answer questions posed in natural language by interpreting the content of an image, with both the image and the question serving as inputs to the model.

The proliferation of satellites and other remote sensing platforms has also enabled more frequent and detailed Earth observation data collection, driving advancements in multitemporal analysis of time-series data \cite{blickensdorfer2024national, liu2024evaluating, valerio2024gee_xtract}. To understand the impact of human activities on surrounding environments and urban development, change detection techniques \cite{yu2024minenetcd, yu2024maskcd, zhang2022multilevel} have been developed to identify changes in land surface dynamics. These techniques generate pixel-wise change masks that classify each pixel as either "changed" or "unchanged." However, such masks often require further interpretation and may not be directly accessible to domain experts unfamiliar with the underlying methods. Inspired by recent NLP advancements, researchers have introduced the concept of remote sensing image change captioning (RSICC) \cite{liu2022remote, hoxha2022change}, a new vision-language task that incorporates image captioning into change detection. This approach aims to provide textual descriptions of detected changes, offering a more intuitive and informative interpretation of change detection results.

Change captioning \cite{park2019robust, hosseinzadeh2021image, kim2021agnostic, guo2022clip4idc} is a recently proposed vision-language downstream task. With the increasing availability of multi-temporal remote sensing data, there is growing interest in using change captioning to study land cover changes and facilitate intelligent interpretation of such images. Automatically generating accurate and detailed captions of land cover changes has important applications in urban planning, disaster management, and environmental monitoring, significantly enhancing our understanding of the evolving built environment \cite{song2014remote, wentz2014supporting}. Change captioning presents some unique challenges compared to traditional image captioning tasks. First, it requires modeling the temporal relationships between bi-temporal images to accurately capture the change of the scene. Additionally, the generated captions must be coherent and contextually accurate throughout the sequence.

RSICC was first introduced by Hoxha et al. \cite{hoxha2022change}, who developed the LEVIR-CCD and DUBAI-CCD datasets and employed CNNs and LSTMs to generate captions. Building on this foundation, Liu et al. \cite{liu2022remote} created the large-scale RSICC dataset and proposed RSICCFormer, a transformer-based model incorporating a Difference Encoding Module and a Multi-Stage Bitemporal Fusion Module. Chang and Ghamisi \cite{chang2023changes} advanced the field with Chg2Cap, which features a dual self-attention transformer encoder and single-layer feature fusion. Cai et al. \cite{cai2023interactive} introduced a model with Multi-Layer Adaptive Fusion and self-gated decoding mechanisms, while Liu et al. \cite{liu2023progressive} proposed PSNet, a pure transformer model leveraging ViT for progressive multi-scale feature fusion.

Recent approaches have shifted towards leveraging large pre-trained models and novel architectures. Liu et al. \cite{liu2023decoupling} introduced a prompt-learning framework that decouples the RSICC task into change detection and identification, utilizing CLIP and fine-tuned frozen language models with learnable prompts. Liu et al. \cite{liu2023pixel} integrated Change Detection pseudo-labels into an auxiliary branch to enhance feature learning and improve caption generation. Liu et al. \cite{liu2024rscama} proposed RSCaMa, a state-space model for joint spatial-temporal modeling, and compared the effectiveness of three different transformer decoder variants within this framework. Yang et al. \cite{yang2024enhancing} introduced an instruction-tuned multimodal framework with a key change feature perception module that jointly optimizes semantic and pixel-level changes, demonstrating the effectiveness of using only key change features as visual instructions for large language models in bi-temporal remote sensing. Wang et al. \cite{wang2024ccexpert} proposed a Difference-aware Integration Module to enhance fine-grained differential features, constructed the extensive CC-Foundation Dataset for remote sensing change captioning, and employed a three-stage training process to optimize the integration of these components with multimodal large language models. These advancements reflect the rapid evolution of methodologies in RSICC. Furthermore, recent works \cite{liu2024change, wang2024changeminds, zhu2024semantic} have combined change detection and change captioning into a unified task, leveraging a single model to address both problems simultaneously.

While significant progress has been made in RSICC, existing transformer-based models often rely on multi-stage fusion of feature maps, which increases complexity and computational cost. This approach tends to overlook the semantic and attribute information of individual objects within feature maps—a critical limitation in remote sensing images characterized by complex backgrounds, high inter-object similarity, and significant variations in object sizes. To address these challenges, we propose a Spatial-Channel attention encoder (SCE) based on a self-attention mechanism. The SCE sequentially models spatial and channel dependencies, capturing relationships between image regions and the semantic attributes of objects. Depth-wise convolution is employed to enhance local information modeling within the Transformer Encoder, while shared encoder parameters improve computational efficiency when processing bi-temporal image pairs. Refined feature maps are concatenated along the channel dimension, fused through a residual block, and passed to the decoder. This design enables the generation of captions that accurately describe specific change regions and object transformations. 

To validate the effectiveness of our proposed method, we conducted extensive experiments on the LEVIR-CC dataset \cite{liu2022remote} and DUBAI-CCD \cite{hoxha2022change}. The results demonstrate that our method achieves competitive performance, with 140.23\% CIDEr on the LEVIR-CC dataset and 97.74\% CIDEr on the DUBAI-CCD dataset. Notably, compared to methods employing multi-stage feature fusion, our approach relying on simple concatenation for feature fusion—achieves superior performance while significantly reducing the number of parameters.

The main contributions of this paper are summarized as follows:
\begin{enumerate}
    \item We analyze the limitations of current transformer-based models for RSICC and highlight the importance of effectively modeling image channels. To address these challenges, we propose a SCE module to better capture relationships between different regions and objects in bi-temporal image pairs.
    \item Compared to approaches that rely on multi-stage feature fusion or computing difference representations for bi-temporal image pairs, our method achieves superior performance using a simpler concatenation strategy for feature fusion.
    \item We conduct a comprehensive set of experiments on the LEVIR-CC and DUBAI-CCD datasets to validate the effectiveness of our proposed method, achieving highly competitive results on both benchmarks.
\end{enumerate}

The rest of this paper is organized as follows. Section II describes the proposed method in detail. Section III presents the information on LEVIR-CC and DUBAI-CCD datasets used in this study and the experimental results. Conclusions and other discussions are summarized in section IV.

\section{Methodology}

\subsection{Overview}

\begin{figure*}
    \centering
    \includegraphics[width=.8\linewidth]{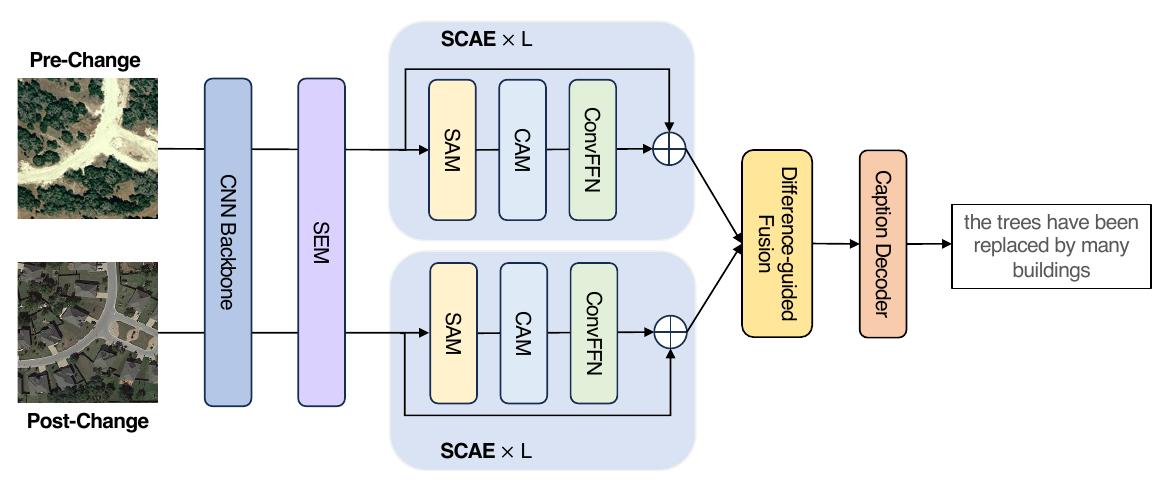}
    \caption{Illustration of SAT-Cap for RSICC.} 
    \label{fig:flowchart}
\end{figure*}

Our proposed SAT-Cap method comprises four main components: a feature extraction module, a spatial-channel attention encoder, a difference-guided fusion module, and a standard transformer-based caption decoder. The model takes as input a bi-temporal image pair along with the corresponding change caption. The overall flow of the model architecture is illustrated in Fig. \ref{fig:flowchart}. 

First, the feature extraction module utilizes a pre-trained ResNet-101 \cite{he2016deep} backbone to extract feature maps from the input image pair. These feature maps are then processed by the spatial-channel attention encoder, which jointly models spatial and channel dependencies. The spatial attention mechanism captures changes across different regions of the image, while the channel attention mechanism extracts rich semantic information about objects within the image.  

After refinement in the encoder, the two feature maps are passed into the fusion module, where they are concatenated along the channel dimension to form a unified visual embedding. This visual embedding is subsequently fed into the transformer decoder, which generates change captions for the bi-temporal image pair.  

\subsection{Image Pair Feature Extraction}

Given an image pair $(I_1, I_2)$, we typically use a pre-trained CNNs to extract features, where $I_i \in \mathbb{R}^{C \times H \times W}$, $i \in (1,2)$ and C, H, W represent the channels, height, and width of the image, respectively. In this paper, we use ResNet-101 \cite{he2016deep} as our backbone to extract feature maps $(X_1, X_2)$. Additionally, we remove the dropout layer and classification layer. Since the channel dimension of the feature maps extracted by CNNs is 2048, to reduce the complexity of the model, we need to decrease the channel dimension. However, if we simply use a convolution layer on the feature maps, it will result in information loss, preventing the model from capturing the semantic information of specific changes in small objects optimally. Therefore, the model finds it more difficult to distinguish between different objects, leading to poorer quality captions when the image pair contains only minor changes.

To learn as precise semantic information as possible, we propose a semantic-enhanced mapping (SEM) module. Compared to simply mapping features extracted by CNNs to a feature space through one convolutional layer, our module can learn richer fine-grained information from the images. Specifically, our module first incorporates positional encoding on the features extracted by the backbone $(X_1, X_2)$, For the input $X_i \in \mathbb{R}^{C^o \times H \times W}$:
\begin{equation}
    X_i^{\prime} = X_i + X_{pos}
\end{equation}
where $i=1$ represents the input before the change, and $i=2$ represents the input after the change. $X_{pos} \in \mathbb{R}^{C^o \times H \times W}$ is a learnable 2D position embedding.

Then we feed the features extracted by the backbone $(X_1^{\prime}, X_2^{\prime})$ into the SEM module, which consists of two standard convolutional layers with Batch Normalization (BN) and Rectified Linear Unit (ReLU) function and one Depthwise convolution layer. The process is given as follows:
\begin{align}
    X_i^{\prime\prime} &= \mathrm{Conv}_2(\mathrm{Conv}_1(X_i^{\prime})) \\
    F_i &= \mathrm{DWConv}(X_i^{\prime\prime})
\end{align}
where $\mathrm{Conv}_i(\cdot)(i = 1,2)$ are two standard $1 \times 1$ convolutional layers with BN and ReLU. $\mathrm{DWConv}(\cdot)$ represents the $3\times3$ depth-wise convolution layer. Through SM module, we obtain a pair of locality-enhanced feature maps, which enhances the model's ability to capture and model small objects..

\subsection{Spatial-Channel Attention Encoder}

\begin{figure}
    \centering
    \includegraphics[width=.8\linewidth]{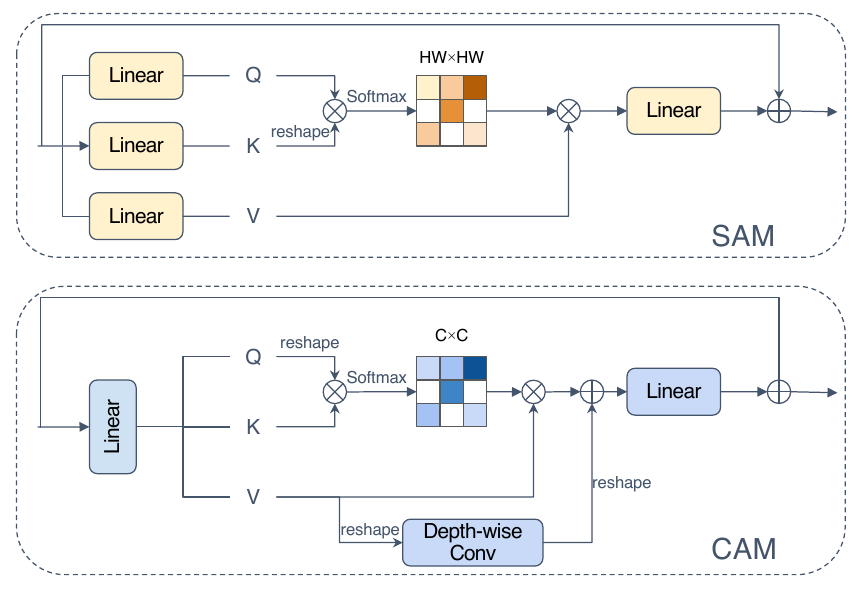}
    \caption{Illustration of SAM and CAM.} 
    \label{fig:SCAE}
\end{figure}

The spatial-channel attention encoder is composed of several stacked Transformer Blocks, each consisting of three parts: Spatial Attention Module (SAM), Channel Attention Module (CAM), and convolutional Feed-Forward Network (convFFN), as shown in Fig. \ref{fig:SCAE}. 

\subsubsection{Spatial Attention Module}

In the SAM, we utilize the Self-Attention Mechanism \cite{vaswani2017attention}, which is widely used in transformer-based methods, to capture the spatial information of the feature map. Given the input $F_i \in \mathbb{R}^{C \times H \times W}$, we first reshape it to $\mathbb{R}^ {HW \times C}$, and then generate query $Q$, key $K$, value $V$ through three linear layers, respectively. This process can be represented as follows:
\begin{equation}
    Q = F_i W_Q, K = F_i W_K, V = F_i W_V 
\end{equation}
where $W_Q, W_K, W_V \in \mathbb{R}^ {C \times C}$ are linear layer learnable parameters. Subsequently, we divide $Q$, $K$, and $V$ into multiple heads $h$, with each head having a dimension $\mathbb{R}^ {C \times \frac{C}{h}}$. Calculate attention separately in each head, then concatenate them together to obtain our spatial feature maps. The entire process is as follows:
\begin{align}
    \text{Head$_i$}(X) &= \text{Softmax}(\frac{Q_i{K_i^\mathrm{T}}}{\sqrt{d_k}}) {V_i} \\
    F_i^{\prime} &= \text{Concat(Head$_1 (F_i)$, $\dots$, Head$_h (F_i)$)}W 
\end{align}
where $\text{Head$_i$}(F_i)$ is the $i$-th head dot-product attention. $W \in \mathbb{R}^{C \times C}$ is the learnable weight matrix to fuse features, $F_i^{\prime}$ is the output feature map.

\subsubsection{Channel Attention Module}

All existing RSICC methods that process feature maps rely only on spatial-based attention mechanisms. These models overlook the relationship along the channel dimension of input feature maps. In previous works \cite{hu2018squeeze, chen2017sca}, it has been shown that different channels often contain different semantic information and attributes. In tasks like change captioning, which involve two input images simultaneously, it's crucial to capture more precise semantic information and attributes. Therefore, CAM is introduced to model the relationship of input feature maps along the channel dimension. Specifically, it generates a global attention map by computing cross-covariance across channels to encode feature maps' semantic information. Here we also employ a self-attention mechanism similar to SAM, with the only difference being that in CAM, this attention calculation is conducted along the channel dimension. Additionally, to enhance the module's ability to understand local contextual features, we directly incorporate depth-wise convolution on the value $V$.

Given the input features $F_i^{\prime} \in \mathbb{R}^{HW\times C}$, our CAM first generates $Q$, $K$, and $V$ through a linear layer. Next, $Q$, $K$, and $V$ are reshaped to their corresponding size $\mathbb{R}^{HW \times C}$ so that channel-wise feature maps can be generated through the dot-product operation of $Q$ and $K$. Meanwhile, we apply depth-wise convolution to $V$ to extract local contextual relationships. The overall process can be summarized as follows:
\begin{align}
    [Q_i^{\prime};K_i^{\prime};V_i^{\prime}] &= F_i^{\prime} W + b  \\
    \text{Attention}(Q_i^{\prime},K_i^{\prime},V_i^{\prime}) &= V_i^{\prime}\cdot \text{Softmax}(\frac{(Q_i^{\prime})^T K_i^{\prime}}{\alpha}) \nonumber\\
    &+ \mathrm{DWConv}(V_i^{\prime}) \\
    F_i^{\prime\prime} &= \text{Attention}(Q_i^{\prime},K_i^{\prime},V_i^{\prime}) + F_i^{\prime}
\end{align}
where $W \in \mathbb{R}^{C \times 3C}$ and $b \in \mathbb{R}^{3C}$ represent the projection weight and bias. $\alpha$ is a learnable parameter to adjust the dot-product. $\mathrm{DWConv}(\cdot)$ represents the $3\times3$ depth-wise convolution layer with BN and GELU non-linear function \cite{devlin2018bert}. $F_i$ and $F_i^{\prime}$ are the input and output features.

\subsubsection{Convolutional Feed-Forward Network}

\begin{figure}
    \centering
    \includegraphics[width=.6\linewidth]{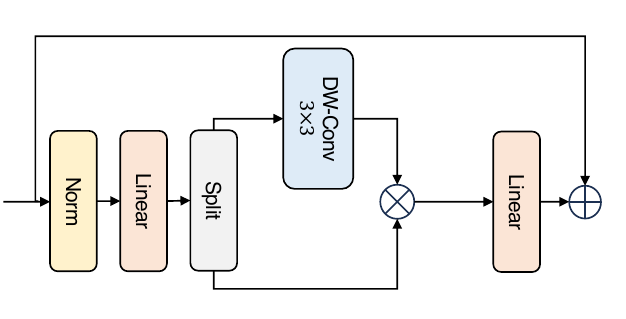}
    \caption{Illustration of ConvFFN.} 
    \label{fig:ConvFFN}
\end{figure}

In the standard transformer model, the FFN typically consists of two linear layers and a non-linear activation function. Since the original transformer architecture was designed for modeling text information, when applied to the computer vision domain, it ignores local information in images. Therefore, in many previous works, DWconv is introduced into the FFN to enhance its capability to model local information:
\begin{equation}
    Z_i = \mathrm{DWConv}(\sigma(F_i^{\prime\prime}W_1+b_1))W_2+b_2
\end{equation}
where $W_1 \in \mathbb{R}^{C \times 4C}$, $W_2 \in \mathbb{R}^{4C \times C}$ and $b_1 \in \mathbb{R}^{4C}$, $b_2 \in \mathbb{R}^{C}$ represent the linear projection weights and biases. $\sigma(\cdot)$ is a non-linear activation function. $\mathrm{DWConv}(\cdot)$ represents the $3\times3$ depth-wise convolution layer. $F_i^{\prime\prime}$ and $Z_i$ are the input and output feature maps. However, adding DWConv significantly increases the parameters of the module. In order to improve the computational efficiency of the module, we first reshape and then divide the feature after the activation function into two features $L_i^1, L_i^2 \in \mathbb{R}^{2C\times H \times W}$ along the channel dimension. Then, we multiply these two features and thereby reducing the parameter count of the second linear layer. ConvMLP is illustrated in Fig. \ref{fig:ConvFFN} and formulated as:
\begin{align}
    [L_i^1; L_i^2] &= \sigma(F_i^{\prime\prime}W_1+b_1)) \\
    \hat{L}_i^1 &= \mathrm{DWConv}(L_i^1) \\
    Z_i &= (\hat{L}_i^1 \odot L_i^2)W_2^{\prime}+b_2^{\prime}
\end{align}
where $W_2^{\prime} \in \mathbb{R}^{2C \times C}$ and $b_2^{\prime} \in \mathbb{R}^{C}$ represent the down-sample projection weight and bias. $\hat{L}_i^1 \in \mathbb{R}^{2C\times H \times W}$ is the feature obtained after $3\times3$ depth-wise convolution layer.

\subsection{Difference-guided Fusion}

After obtaining the feature maps refined by spatial-channel attention encoder for bi-temporal image pairs, we need to design a fusion block to integrate the information of the two feature maps, thereby guiding the caption decoder to generate higher-quality text. Inspired by Change2Captions \cite{chang2023changes}, we first calculate the cosine similarity of two feature maps. Then, we add it to feature maps and concatenate them together. As the dimension of the resulting feature increases, to reduce the parameter and improve computational efficiency, we pass it through a convolutional layer for dimensionality reduction. Finally, the dimension-reduced feature is passed into a Resblock to obtain more accurate feature. This can be expressed as:
\begin{align}
    Sim &= \text{Cos}(Z_1, Z_2) \\
    F_f &= [Z_1 + Sim; Z_2 + Sim] \\
    F_f^{\prime} &= \mathrm{Conv}_3(\mathrm{Conv}_2(\mathrm{Conv}_1(F_f)))+F_f
\end{align}
where Cos$(\cdot, \cdot)$ can calculate the cosine similarity between two tensors. $\mathrm{Conv}_i(\cdot)(i = 1,2,3)$ are three convolutional layers with BN and ReLU, their kernel sizeas are $1\times1$, $3\times3$, and $1\times1$. After obtaining the fused feature $F_f^{\prime} \in \mathbb{R}^{C\times H \times W}$, we reshape it and feed it into the decoder of the model to guide the decoder in generating text describing the changes.

\subsection{Caption Decoder}

To generate text describing changes in remote sensing image pairs, we use a widely adopted transformer-based decoder. It consists of a series of stacked transformer decoder blocks, where each transformer decoder block comprises a masked multiheaded self attention mechanism, an encoder-decoder cross-attention mechanism, and a feed-forward network. The calculation within a transformer decoder block can be represented as follows:
\begin{align}
    X_{l}^{\prime} &= \text{LN}(X_{l-1} + \text{M-MSA}(X_{l-1})) \\
    X_{l}^{\prime\prime} &= \text{LN}(X_{l}^{\prime} + \text{CA}(X_{l}^{\prime}, F_f^{\prime}, F_f^{\prime})) \\
    X_{l} &= \text{LN}(X_{l}^{\prime\prime} + \text{MLP}(X_{l}^{\prime\prime}))
\end{align}
where $X_{l-1}$ and $X_{l}$ denote the input and output of the l-th transformer decoder block, M-MSA, CA, MLP and LN denote the masked multiheaded self attention, encoder-decoder cross-attention, feed-forward network, and layer normalization, respectively.

\section{Experiments}

\subsection{Dataset Descriptions}

We adopt two benchmark datasets to evaluate the performance of the proposed method on RSICC tasks.

\subsubsection{\textbf{LEVIR-CC Dataset}}

The images in this dataset \cite{liu2022remote} are mainly collected from the LEVIR-CD dataset \cite{chen2020spatial}, which consists of images from 20 regions in Texas, United States. Each image has dimensions of 1024 by 1024 pixels, with a spatial resolution of 0.5 meters. The time span for each change image pair ranges from 5 to 15 years. And LEVIR-CC comprises a total of 10,077 pairs of images, each with dimensions of $256 \times 256$ pixels. Additionally, each image pair is accompanied by five corresponding texts describing whether changes have occurred. The maximum length of the text is 39, and the average length of text is 7.99. In total, LEVIR-CC contains 5038 pairs of images with changes and 5039 pairs without changes. We use a pre-defined partition of the dataset to validate the effectiveness of the model, with 6815 image pairs used for training, 1333 image pairs for validation, and the remaining 1929 iamge pairs for testing.

\subsubsection{\textbf{DUBAI-CCD Dataset}}

This dataset \cite{hoxha2022change} describes the urban changes in Dubai between May 19, 2000, and June 16, 2010, and includes a total of 500 bi-temporal image pairs, each with dimensions of 50 × 50 pixels. Each image pair is accompanied by 5 related change description captions, with a maximum length of 23 words and an average length of 7.35 words. We use the pre-divided portions of the dataset, with 300 image pairs serving as the training set, 50 image pairs as the validation set, and the remaining 150 image pairs as the test set to evaluate the model's effectiveness.

\subsection{Experimental Settings and Implementation Details}

In our experiments, we set the maximum number of iterations to 50 epochs for both LEVIR-CC dataset and DUBAI-CCD dataset. The Adam optimizer \cite{kingma2014adam} with learning rate of $1e-4$ is employed during the training phase. We use a batch size of 32 for LEVIR-CC dataset and a batch size of 8 for DUBAI-CCD dataset. We use a pre-trained ResNet-101 to extract feature maps from the image pairs. The feature dimensions in both the transformer encoder and transformer decoder are set to 512, and the hidden dimension in the feed-forward layer is set to 2048, and the number of self-attention heads is 8. Through experimental validation, we set the number of layers for transformer encoder to 3, the number of layers for transformer decoder to 1. Since ResNet-101 is pre-trained on natural image datasets, to better adapt it to the RSICC task, we first fine-tune ResNet-101 on the current datasets and then use the fine-tuned model as our backbone. The experiment is conducted on an Nvidia RTX A6000 GPU.

Following the standard evaluation metrics in previous works \cite{liu2022remote, chang2023changes}, BLEU-N (where N=1,2,3,4) \cite{papineni2002bleu}, ROUGE-L \cite{lin2004rouge}, METEOR \cite{banerjee2005meteor}, and CIDEr-D \cite{vedantam2015cider} are adopted to evaluate our proposed method and compared with other SOTA methods.

\subsection{Performance Comparison to the State of the Art}

\begin{table*}
\centering
\caption{Comparison with the state of the art on the LEVIR-CC dataset. All values are reported as percentage (\%), and the best results are highlighted in bold.}
\label{tab:Comarison}
\resizebox{\textwidth}{!}{
\begin{tabular}{l|ccccccc|c}
\toprule
Method & BLEU-1 &    BLEU-2 &    BLEU-3 &    BLEU-4 &    METEOR &    ROUGE-L &   CIDEr-D & S$_m^{\star}$ \\
\midrule
DUDA \cite{park2019robust} & 81.44 & 72.22 & 64.24 & 57.79 &37.15 &71.04 &124.32 & 72.58 \\
MCCFormer-S \cite{qiu2021describing} &79.90 &70.26 &62.68 &56.68 &36.17 &69.46 &120.39 & 70.68\\
MCCFormer-D \cite{qiu2021describing} & 80.42 &70.87 &62.86 &56.38 &37.29 &70.32 &124.44 & 72.11\\
RSICCFormer \cite{liu2022remote} &    84.72 & 76.27 & 68.87 & 62.77 & 39.61 & 74.12 & 134.12 &  77.65 \\
Chg2Cap \cite{chang2023changes} &   \textbf{86.14} & 78.08 & 70.66 & 64.39 & 40.03 & 75.12 & 136.61 & 79.04 \\ 
\midrule
Ours & \textbf{86.14} & \textbf{78.19} & \textbf{71.44} & \textbf{65.82} & \textbf{40.51} & \textbf{75.37} & \textbf{140.23} & \textbf{80.48} \\
\bottomrule
\end{tabular}
}
\end{table*}

\begin{table*}
\centering
\caption{Comparison with the state of the art on the DUBAI-CCD dataset. All values are reported as percentage (\%), and the best results are highlighted in bold.}
\label{tab:Comarison_dubai}
\resizebox{\textwidth}{!}{
\begin{tabular}{l|ccccccc|c}
\toprule
Method & BLEU-1 &    BLEU-2 &    BLEU-3 &    BLEU-4 &    METEOR &    ROUGE-L &   CIDEr-D & S$_m^{\star}$  \\ 
\midrule
DUDA \cite{park2019robust} &58.82 &43.59 &33.63 &25.39 &22.05 &48.34 &62.78 &39.64\\
MCCFormer-S \cite{qiu2021describing} &52.97 &37.02 &27.62 &22.57 &18.64 &43.29 &53.81 &34.58\\
MCCFormer-D \cite{qiu2021describing} &64.65 &50.45 &39.36 &29.48 &25.09 &51.27 &66.51 &43.09\\
RSICCFormer \cite{liu2022remote} &67.92 &53.61 &41.37 &31.28 &25.41 &51.96 &66.54 &43.80\\
Chg2Cap \cite{chang2023changes} &   72.04 &60.18 &\textbf{50.84} &\textbf{41.70} &28.92 &58.66 &92.49 & 55.44\\ 
\midrule
Ours &\textbf{73.48}  &\textbf{60.98}  &50.51 &40.80  &\textbf{29.62}  &\textbf{59.06}   &\textbf{97.74}  & \textbf{56.81}\\
\bottomrule
\end{tabular}
}
\end{table*}

In this section, we compare our proposed method with the current Change Captioning methods on the LEVIR-CC dataset, including three methods for natural image change captioning, DUDA \cite{park2019robust}, MCCFormer-S \cite{qiu2021describing}, MCCFormer-D \cite{qiu2021describing},  and three methods for remote sensing image change captioning, RSICCFormer \cite{liu2022remote}, Chg2Cap \cite{chang2023changes}, RSCaMa \cite{liu2024rscama}.

We compare the performance of our method with the aforementioned methods in Table \ref{tab:Comarison}. It can be observed that for the LEVIR-CC dataset, our method achieves SOTA performance across all metrics. Specifically, our method outperforms the RSCaMa method by 0.58\% in BLEU-4 score and the Chg2Cap method by 0.48\% in METEOR score. Additionally, and most notably, our method achieves a CIDEr-D score of 140.23\%, representing an improvement of 3.62\% over the Chg2Cap method.

Additionally, we present the performance of different methods on the DUBAI-CCD dataset in Table \ref{tab:Comarison_dubai}. Our method achieves SOTA performance across most metrics. Similar to the results on the LEVIR-CC dataset, our method shows a significant improvement of 5.25\% in CIDEr-D compared to the Chg2Cap method. 

Overall, the results indicate that our method consistently performs well on both datasets. This not only suggests that simple fusion methods can be highly effective for transformer-based encoder-decoder models, but also underscores the importance of joint spatial and channel modeling in change captioning tasks. Joint modeling enables the extraction of features and semantic attributes from different regions and objects, thereby better capturing changes in bi-temporal image pairs and guiding the decoder to generate more accurate captions.

\subsection{Ablation Studies}

We conduct a comprehensive series of experiments on the LEVIR-CC dataset to validate the effectiveness of the components proposed in our method. Additionally, we explore the impact of various factors on the results, including different dimensions for computing Cosine Similarity, different fusion methods, and different feature map spatial sizes. In the ablation studies, all experimental settings remain consistent with those described in the previous experiments.

\subsubsection{Effectiveness of different Components}

Table \ref{tab:Ablation} shows the impact of the SEM and ConvFFN modules on the model's performance. We can observe that the inclusion of SEM and ConvFFN significantly enhances the model's performance, highlighting the importance of these two modules. Specifically, compared to the method without SEM and ConvFFN, the simultaneous addition of these modules achieves an improvement of 1.69\% in BLEU-4 and 2.17\% in CIDEr-D, with only a minimal increase in parameters (0.05 M).

\begin{table*}
\centering
\caption{Effectiveness of SM and convFFN on the LEVIR-CC dataset. All values are reported as percentage (\%), and the best results are highlighted in bold.}
\label{tab:Ablation}
\resizebox{\textwidth}{!}{
\begin{tabular}{cc|ccccccc|c}
\toprule
SEM  & ConvFFN & BLEU-1 &    BLEU-2 &    BLEU-3 &    BLEU-4 &    METEOR &    ROUGE-L &   CIDEr-D & Param \\ 
\midrule
\XSolidBrush &\XSolidBrush &85.37  &76.97  &69.96 &64.13 &40.33  &74.77   &138.06 & 20.35M \\
\Checkmark  &\XSolidBrush &85.62  &77.21  &69.92  &63.80  &40.41  &75.12   &139.39 &21.94M\\
\XSolidBrush &\Checkmark &85.70  &77.54  &70.40  &64.70  &39.90  &74.82   &136.88 &18.82M \\
\Checkmark &\Checkmark & \textbf{86.14} & \textbf{78.19} & \textbf{71.44} & \textbf{65.82} & \textbf{40.51} & \textbf{75.37} & \textbf{140.23} &20.40M \\
\bottomrule
\end{tabular}
}
\end{table*}

\subsubsection{Impact of CosineSimilarity along different dimensions}

Inspired by Chg2Cap \cite{chang2023changes}, we calculate the cosine similarity of the two image features before fusing the bi-temporal image pair features and add it to the respective image features. However, Chg2Cap did not discuss the dimensions along which the cosine similarity was calculated. Therefore, in this part, we validate the impact of calculating cosine similarity based on different dimensions on performance. First, we list the experimental results without calculating cosine similarity in Table \ref{tab:cos}. Then, we present the results of experiments calculating cosine similarity along the individual dimensions of height, width, channel, as well as various combinations of these dimensions. We observe that, compared to calculating the cosine similarity along the height and width dimensions, calculating it solely along the channel dimension achieves better results. Therefore, in our final method, we chose to calculate cosine similarity only along the channel dimension.

\begin{table*}
\centering
\caption{Impact of adding CosineSimilarity along different dimensions on the LEVIR-CC dataset. All values are reported as percentage (\%), and the best results are highlighted in bold.}
\label{tab:cos}
\resizebox{\linewidth}{!}{
\begin{tabular}{l|ccccccc}
\toprule
CosineSimilarity dim & BLEU-1 &    BLEU-2 &    BLEU-3 &    BLEU-4 &    METEOR &    ROUGE-L &   CIDEr-D \\ 
\midrule
w/o CosineSimilarity &84.97  &76.65  &69.64  &63.85  &39.71  &74.01 &136.07\\
\midrule
along height dim &86.09  &77.90  &70.96  &65.23  &40.45  &75.41 & 139.87\\
along width dim &86.02  &77.81  &70.89  &65.26  &40.39  &75.21   &139.80\\
along height$\times$width dim &86.02  &77.65  &70.65  &64.99  &40.34  &75.30   &139.67 \\
along height and width dims &85.99  &77.94  &71.08  &65.55  &40.12  &75.03   &139.75\\
along channel dim & \textbf{86.14} & \textbf{78.19} & \textbf{71.44} & \textbf{65.82} & \textbf{40.51} & 75.37 & 140.23\\
along channel, height$\times$width dims & 85.73  & 77.46  & 70.58  & 64.94  & 40.16  & 74.86   & 138.83 \\
along channel, height, and width dims &85.93  &77.81  &71.01  &65.47  &40.46  &\textbf{75.48}   &\textbf{140.53} \\
\bottomrule
\end{tabular}}
\end{table*}

\subsubsection{Impact of different fusion methods in Difference-guided Fusion}

In Table \ref{tab:fusion}, we compare four different methods for fusing bi-temporal image pair features. The experimental results indicate that concatenating the features of the two images yields the best performance. By concatenating the image features and then adding a $1\times1$ convolutional layer, we can rearrange the concatenated features to form a new feature, thereby minimizing the semantic information loss of the two features refined by the transformer encoder.

\begin{table}
\centering
\caption{Impact of different fusion methods in Difference-guided Fusion module on the LEVIR-CC dataset. All values are reported as percentage (\%), and the best results are highlighted in bold.}
\label{tab:fusion}
\resizebox{.6\linewidth}{!}{
\begin{tabular}{l|ccccccc}
\toprule
Fusion methods & B1 &    B4 &    M &    R &   C \\ 
\midrule
Sub &86.07    &64.15  &40.42  &75.36   &138.46\\
Sum &85.26    &65.31  &39.86  &74.92   &138.07\\
Element-wise Product &84.96    &64.53  &39.99  &74.70   &138.07\\
Concatenate & \textbf{86.14}  & \textbf{65.82} & \textbf{40.51} & \textbf{75.37} & \textbf{140.23}\\
\bottomrule
\end{tabular}}
\end{table}

\subsubsection{Impact of different feature map size}

In table \ref{tab:spatial_size}, we validate the impact of different feature map spatial sizes on model performance. The feature maps extracted by ResNet-101 have a spatial size of 8 $\times$ 8. Subsequently, we use average pooling to set the spatial size to 7 $\times$ 7 and 9 $\times$ 9. The results indicate that the model already achieves optimal performance with the default spatial size of 8 $\times$ 8.

\begin{table}
\centering
\caption{Impact of different feature map size on the LEVIR-CC dataset. All values are reported as percentage (\%), and the best results are highlighted in bold.}
\label{tab:spatial_size}
\resizebox{.6\linewidth}{!}{
\begin{tabular}{c|ccccccc}
\toprule
Feature map Size & B1 & B4 &    M &    R &   C \\ 
\midrule
7 $\times$ 7 &85.19   &63.25  &40.33  &75.14   &138.94\\
8 $\times$ 8 & \textbf{86.14}  & \textbf{65.82} & \textbf{40.51} & \textbf{75.37} & \textbf{140.23}\\
9 $\times$ 9 &86.05 &64.44  &40.37  &75.15   &139.08\\
\bottomrule
\end{tabular}}
\end{table}

\subsubsection{Parametric Analysis}

In a transformer-based encoder-decoder architecture, the number of layers in the encoder and decoder is a crucial hyper-parameter. It significantly impacts the model size, computational cost, and overall performance. To further verify whether we have selected the appropriate number of layers for the encoder and decoder, we present a performance comparison of different transformer encoder and decoder layer configurations in Table \ref{tab:depth}. It can be observed that as the number of transformer encoder layers increases, the model's performance gradually improves. However, with the same number of encoder layers, the model's performance decreases as the number of decoder layers increases. This decline in performance may be due to the increased difficulty in optimizing the model with more decoder layers, making it more challenging to generate high-quality captions. Therefore, we ultimately chose a model with three transformer encoder layers and one transformer decoder layer.

\begin{table}
\centering
\caption{Performance Comparisons with different depth of encoder and decoder on the LEVIR-CC Dataset. All values are reported as percentage (\%) except for parameter, and the best results are highlighted in bold.}
\label{tab:depth}
\resizebox{.6\linewidth}{!}{
\begin{tabular}{cc|ccccccc|c}
\toprule
E & D & B1 &    B4 &    M &    R &   C & Param\\ 
\midrule
1&1 &85.56    &64.65  &39.29  &74.39   &135.18 & 13.01M\\
2&1 &84.69    &64.86  &39.95  &74.75   &136.55& 16.70M\\
3&1 &\textbf{86.14}  & \textbf{65.82} & \textbf{40.51} & \textbf{75.37} & \textbf{140.23} & 20.40M\\
3&2 &85.93    &64.82  &\textbf{40.51}  &75.21   &138.96 &25.13M\\
3&3 &85.51    &64.30  &40.09  &74.91   &138.37 &29.86M\\
\bottomrule
\end{tabular}}
\end{table}

\subsubsection{Impact of different local enhance methods in channel attention module}

To extract both global and local information from the images, we introduce a local enhancement branch in the channel attention module. In Table \ref{tab:channel_enhance}, we compare different local enhancement methods. Compared to other methods, incorporating a $3 \times 3$ DWConv achieved the best results across most metrics.

\begin{table}
\centering
\caption{Impact of different local enhance methods in channel attention module on the LEVIR-CC dataset. All values are reported as percentage (\%), and the best results are highlighted in bold.}
\label{tab:channel_enhance}
\resizebox{.6\linewidth}{!}{
\begin{tabular}{l|ccccccc}
\toprule
Methods & B1 & B4 &    M &    R &   C \\ 
\midrule
SE \cite{hu2018squeeze} (r=4) &85.45    &63.56  &40.38  &\textbf{75.59}  &139.95\\
SE \cite{hu2018squeeze} (r=8) &85.59    &64.23  &39.66  &75.16  &138.60 \\
SE \cite{hu2018squeeze} (r=16) &85.94    &.64.47  &40.39  &75.45   &140.11 \\
SE \cite{hu2018squeeze} (r=32) &85.62    &63.64  &39.09  &74.24 &134.09 \\
1 $\times$ 1 Conv &86.06    &64.66  &40.47  &75.23  & 138.56 \\
3 $\times$ 3 Conv  &85.72    &64.30  &40.30  &75.13  &138.84 \\
\midrule
DWConv & \textbf{86.14}  & \textbf{65.82} & \textbf{40.51} & 75.37 & \textbf{140.23} \\
\bottomrule
\end{tabular}}
\end{table}

\subsubsection{Effect of different combinations of spatial-attention module and channel-attention module}

Since SAM and CAM model the spatial and channel aspects respectively, their combination sequence significantly affects the model's performance. Therefore, in this part, we primarily investigate three different combination methods of SAM and CAM: first passing through SAM then CAM (SAM $\to$ CAM), first passing through CAM then SAM (CAM $\to$ SAM), and passing through SAM and CAM in parallel (SAM + CAM).

Table \ref{tab:combination} lists the performance of the three different combination methods. To validate the importance of joint spatial and channel modeling, the performance of using only SAM or CAM is also included. Compared to modeling only the spatial or channel aspects, all three combination methods that integrate both show improved performance. This indicates that both spatial attention and channel attention are crucial for the change captioning task. Moreover, among the three combination methods, the approach of first passing through SAM and then through CAM performs the best.

\begin{table}
\centering
\caption{Performance comparison of different combinations of SAM and CAM on the LEVIR-CC dataset. All values are reported as percentage (\%), and the best results are highlighted in bold.}
\label{tab:combination}
\resizebox{.6\linewidth}{!}{
\begin{tabular}{l|ccccccc}
\toprule
Combinations & B1 &    B4 &    M &    R &   C \\ 
\midrule
SAM &85.82   &64.53  &40.39  &74.99  &138.67\\
CAM &85.85    &64.30  &39.90  &74.31 &136.42\\
SAM $\to$ CAM & \textbf{86.14}  & \textbf{65.82} & \textbf{40.51} & \textbf{75.37} & \textbf{140.23}\\
CAM $\to$ SAM &85.51  &65.12  &40.30  &75.06   &139.39\\
SAM + CAM &85.70  &65.26  &40.22  &75.14   &139.10\\
\bottomrule
\end{tabular}}
\end{table}

\subsubsection{Effect of Layernorm weight sharing in spatial-channel attention encoder}

In Table \ref{tab:layernorm}, we list the impact of using shared layernorm on model performance in two different combination scenarios. The results indicate that employing shared layernorm for SAM and CAM in the spatial-channel attention encoder under different conditions can effectively enhance model performance while reducing certain model parameters.

\begin{table}
\centering
\caption{Effect of Layer norm  weight sharing in spatial-channel attention encoder on the LEVIR-CC Dataset. All values are reported as percentage (\%), and the best results are highlighted in bold.}
\label{tab:layernorm}
\resizebox{.6\linewidth}{!}{
\begin{tabular}{l|ccccccc}
\toprule
Combinations & B1 &    B4 &    M &    R &   C \\ 
\midrule
CAM $\to$ SAM (w/o) & \textbf{85.55}    &64.88  &40.11  &74.93  &138.74\\
CAM $\to$ SAM (w/ ) &85.51    & \textbf{65.12}  & \textbf{40.30}  & \textbf{75.06}   & \textbf{139.39} \\
\midrule
SAM $\to$ CAM (w/o) & 86.03   &65.06  &40.43  &75.30  &139.91\\
SAM $\to$ CAM (w/ ) & \textbf{86.14}  & \textbf{65.82} & \textbf{40.51} & \textbf{75.37} & \textbf{140.23}\\
\bottomrule
\end{tabular}}
\end{table}

\subsection{Further Comparisons}

Following \cite{liu2022remote}, we also employ the same three different aspects to evaluate the effectiveness of our method. These three aspects are: only test bi-temporal image pairs with no changes, only test bi-temporal image pairs with changes, and test the entire test set. The final results are reported in Table \ref{tab:three}. 

It can be observed that for unchanged remote sensing images, the performance of Chg2Cap surpasses our method. This may be attributed to the introduction of SEM and CAM, making the model more sensitive to changes in small objects. Due to the resolution of the LEVIR-CC dataset being only $256\times256$, this can lead to some misjudgments by the model. However, it is precisely due to the inclusion of these two modules that our method significantly outperforms RSICCFormer and Chg2Cap in describing changed remote sensing image pairs. This further validates the importance of joint spatial-channel modeling. It not only captures changes in different regions of image pairs but also captures changes in different objects, thereby enhancing the model's ability to model image pairs and generate a more accurate image feature.

\begin{table*}
\centering
\caption{Performance Comparison under three different aspects on the LEVIR-CC Dataset. All values are reported as percentage (\%), and the best results are highlighted in bold.}
\label{tab:three}
\resizebox{\linewidth}{!}{
\begin{tabular}{c|l|ccccccc}
\toprule
Test Set & Method & BLEU-1 &    BLEU-2 &    BLEU-3 &    BLEU-4 &    METEOR & ROUGE-L &   CIDEr-D \\ 
\midrule
\multirow{3}{*}{only no-change}& RSICCFormer \cite{liu2022remote} &95.05 &94.24 &93.76 &93.42 &72.20 &95.68 & - \\
& Chg2Cap \cite{chang2023changes} &\textbf{98.08}  &\textbf{97.83}  &\textbf{97.70}  &\textbf{97.61}  &\textbf{78.21}  &\textbf{98.22}  & - \\
& Ours &97.80  &97.44  &97.33  &97.29  &76.23  &97.49   & -
\\
\midrule
\multirow{3}{*}{only change}& RSICCFormer \cite{liu2022remote} &76.43 &61.92 &48.81 &38.14 &25.72 &52.53 &60.56 \\
& Chg2Cap \cite{chang2023changes} &\textbf{77.32}  &\textbf{63.26}  &50.06  &39.09  &25.73  &52.02   &58.30 \\
& Ours &77.12  &63.20  &\textbf{51.20}  &\textbf{41.46}  &\textbf{26.28}  &\textbf{53.23}   &\textbf{68.91}\\
\midrule
\multirow{3}{*}{entire set}& RSICCFormer \cite{liu2022remote} &    84.72 & 76.27 & 68.87 & 62.77 & 39.61 & 74.12 & 134.12\\
& Chg2Cap \cite{chang2023changes} &\textbf{86.14}  &78.04  &70.66  &64.39  &40.03  &75.12   &136.61 \\
& Ours & \textbf{86.14} & \textbf{78.19} & \textbf{71.44} & \textbf{65.82} & \textbf{40.51} & \textbf{75.37} & \textbf{140.23}\\
\bottomrule
\end{tabular}}
\end{table*}

\subsection{Visualization}

\begin{figure}
    \centering
    \includegraphics[width=.6\linewidth]{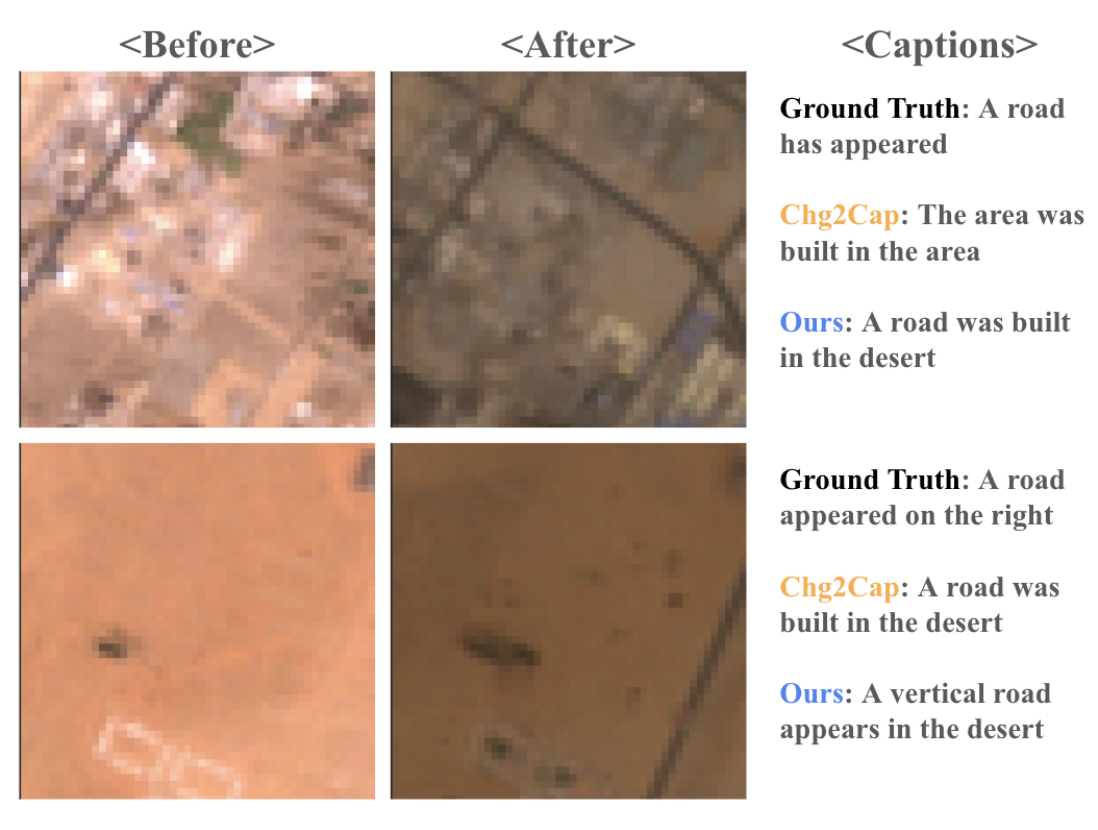}
    \caption{Captioning results on the DUBAI-CCD dataset. Black sentence is one of the five ground truth sentences. The orange captions are generated by Chg2Cap \cite{chang2023changes}, while the captions generated by our method are shown in blue.}
    \label{fig:visual_1}
\end{figure}

\begin{figure}
    \centering
    \includegraphics[width=.6\linewidth]{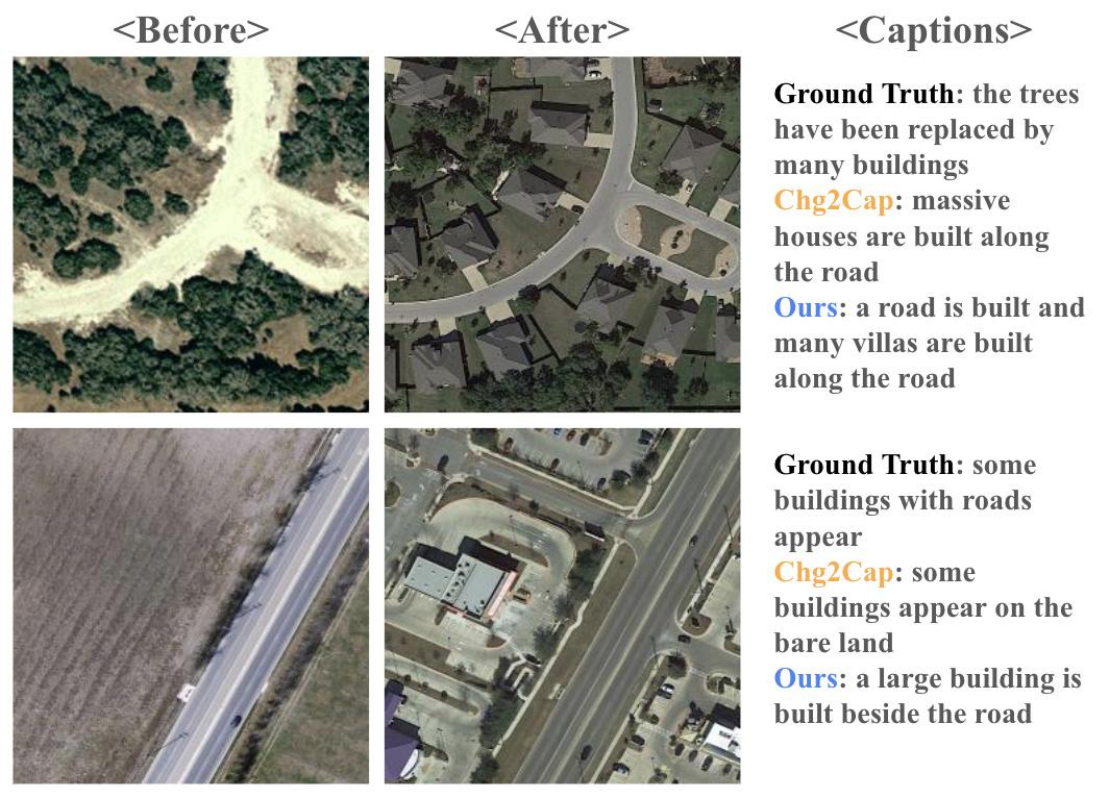}
    \caption{Captioning results on the LEVIR-CC dataset. Black sentence is one of the five ground truth sentences. The orange captions are generated by Chg2Cap \cite{chang2023changes}, while the captions generated by our method are shown in blue.}
    \label{fig:visual_2}
\end{figure}

In Fig. \ref{fig:visual_1} and Fig. \ref{fig:visual_2}, we visualize two different examples from the LEVIR-CC dataset and the DUBAI-CCD dataset, respectively, to compare the groud truth captions, the captions generated by Chg2Cap \cite{chang2023changes} and the captions generated by our method. 

In the first example shown in Fig. \ref{fig:visual_1}, the caption generated by Chg2Cap is entirely unusable, whereas our method successfully identifies the construction of a road in the area and accurately determines that the background of the image pair is a desert environment. This demonstrates that by employing joint spatial-channel modeling, our model effectively focuses on the regions of change and captures the semantic information and attributes of the image pair. Furthermore, in the second example, our model not only identifies the road and the desert but also recognizes that the road is vertical, indicating that the model has learned richer semantic information and attributes of the road.

Furthermore, we can observe from Fig. \ref{fig:visual_2} that in the first bi-temporal image pair, Chg2Cap successfully expressed
that many houses were built along the road, but it lacked a
description of the road itself. Our method not only depicted
that villas were built along the road, but also described that the
road was newly built. In the second bi-temporal image pair,
the caption generated by our method not only successfully
described that the building was built by the roadside, but also
emphasized that it is a large building. This indicates that after
joint spatial-channel modeling, our method effectively focuses
on different regions of the image while also paying attention
to the features and attributes of different objects in the image.
This further validates the effectiveness of our approach.

\begin{figure}
    \centering
    \includegraphics[width=.6\linewidth]{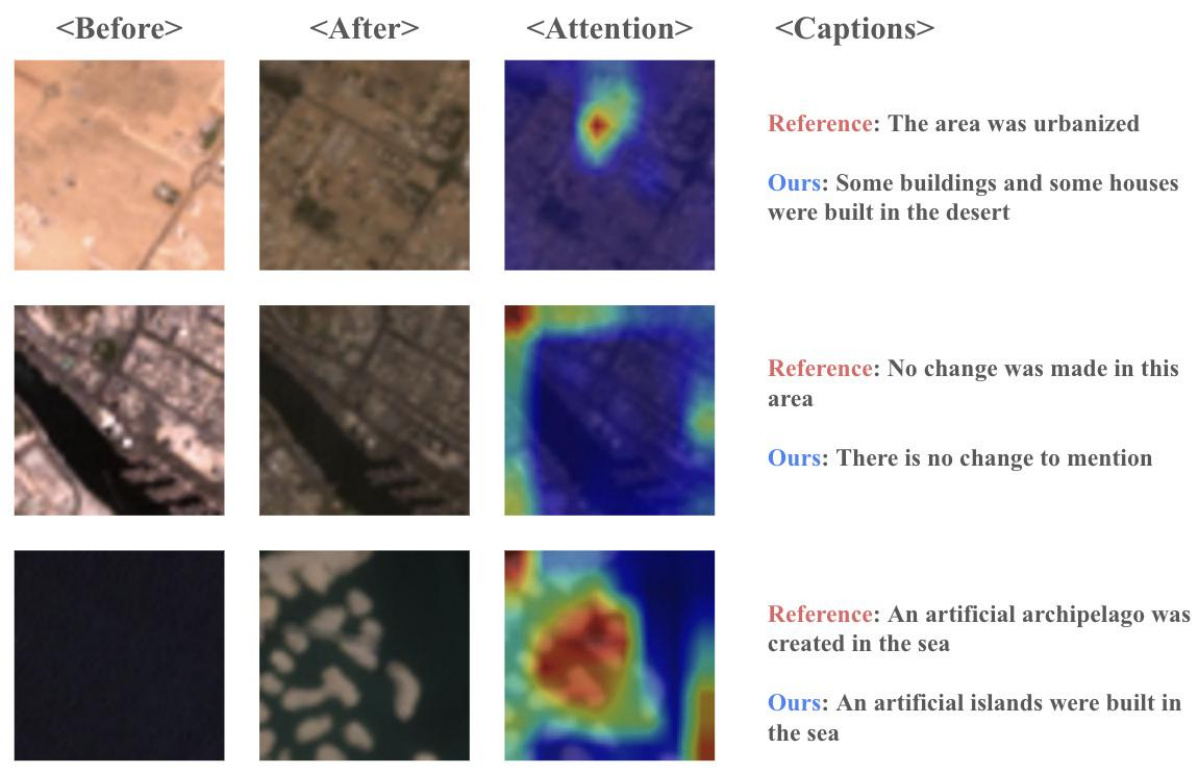}
    \caption{Typical captioning results on the DUBAI-CCD dataset. From left to right, the figure includes the pre-change image, post-change image, attention map over post-change image, and captions from ground truth and generated by our method.}
    \label{fig:visual_3}
\end{figure}

\begin{figure}
    \centering
    \includegraphics[width=.6\linewidth]{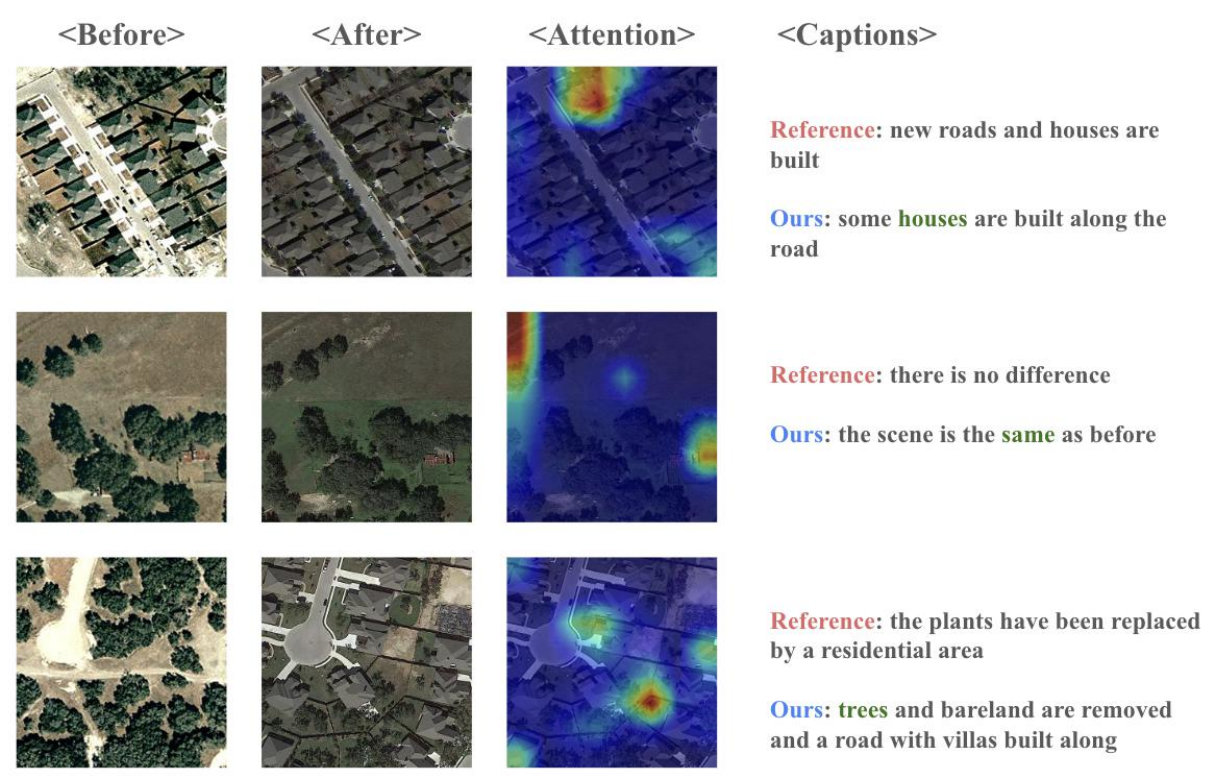}
    \caption{Typical captioning results on the LEVIR-CC dataset. From left to right, the figure includes the pre-change image, post-change image, attention map over post-change image, and captions from ground truth and generated by our method.}
    \label{fig:visual_4}
\end{figure}

In Fig. \ref{fig:visual_3} and Fig. \ref{fig:visual_4}, we visualize three examples from the LEVIR-CC dataset and the DUBAI-CCD dataset, respectively, to further compare the reference captions and the captions generated by our method. In these figures, "Before" denotes the image before the change, "After" denotes the image after the change, "Attention" represents the attention map over the image in the transformer decoder, "Reference" in the captions refers to the ground truth (GT), and "Ours" refers to the captions generated by our method. 

In the DUBAI-CCD dataset, the spatial resolution of the change image pairs is 30 meters, making many changes difficult to identify, particularly for features such as roads and houses. However, as shown in the first example of Fig. \ref{fig:visual_3}, the caption generated by our method, compared to the reference, not only depicts the urbanization of the area but also provides a more detailed description by identifying many buildings and houses being constructed in the desert. This demonstrates that our joint spatial-channel approach can recognize the desert background of the image and also identify small objects like buildings and houses, as further validated by the attention map. In the second example, our caption accurately identifies that there is no change in the image pair. In the last example, the attention map shows that our model successfully focuses on the island part of the image and accurately describes the changes in that area. The examples from the LEVIR-CC dataset, as shown in Fig. \ref{fig:visual_4}, also demonstrate that our model can accurately focus its attention on specific target objects, such as "houses" and "trees". 

\begin{figure*}
    \centering
    \includegraphics[width=\textwidth]{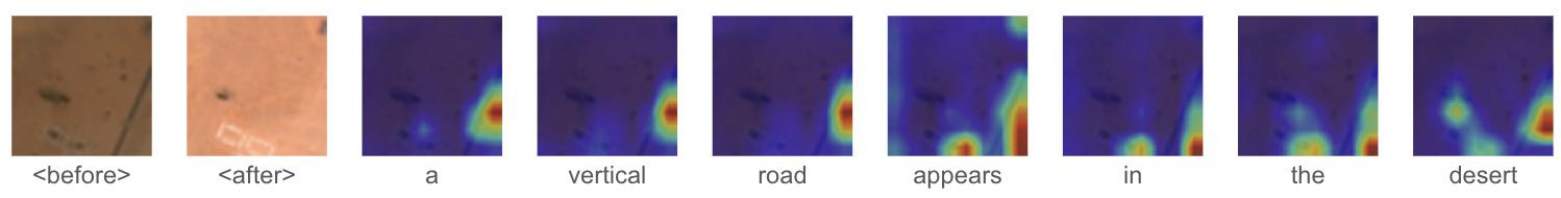}
    \caption{The visualization of attended images along with the caption generation for processes for our method on the DUBAI-CCD dataset.}
    \label{fig:visual_5}
\end{figure*}

\begin{figure}
    \centering
    \includegraphics[width=.8\linewidth]{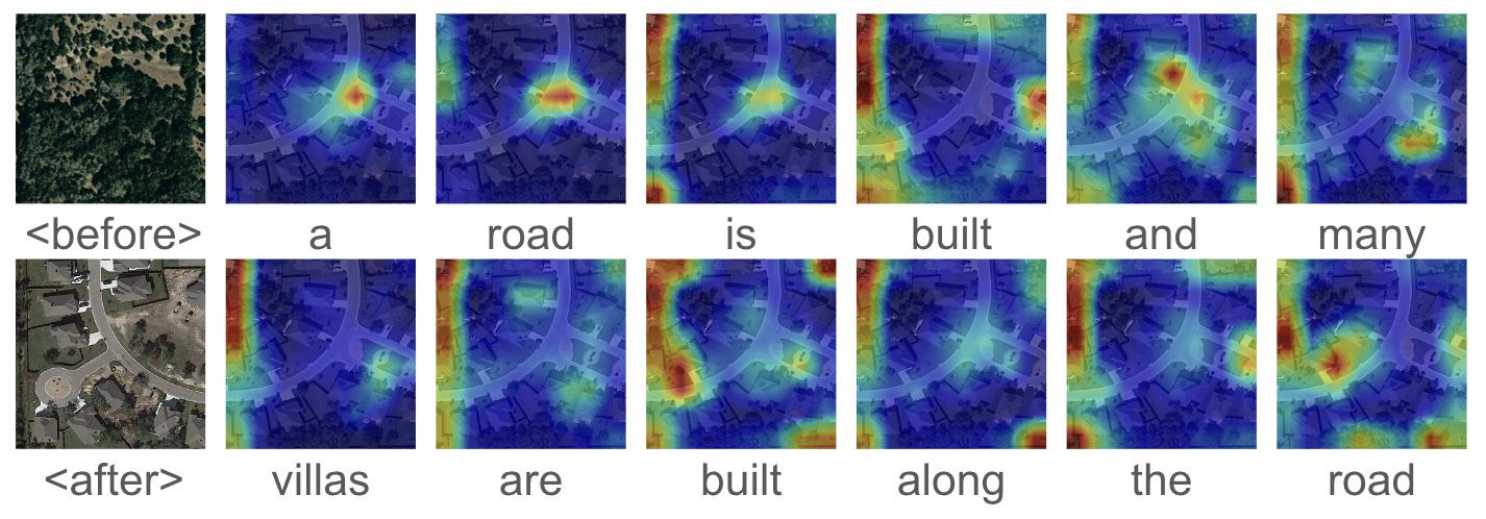}
    \caption{The visualization of attended images along with the caption generation for processes for our method on the LEVIR-CC dataset.}
    \label{fig:visual_6}
\end{figure}

To better qualitatively evaluate the captions generated by our method, we visualized each generated word and its corresponding attended images in Fig. \ref{fig:visual_3} and Fig. \ref{fig:visual_4}. From Fig. \ref{fig:visual_3}, it can be observed that our proposed method effectively focuses attention on objects such as roads and villas, enabling the generation of more detailed descriptions. Notably, in Fig. \ref{fig:visual_4}, compared to the word "road", the attended image for the word "desert" broadens the scope of attention, thereby better identifying the background category of the image. These examples not only verify that our model can accurately recognize changes in different regions of the image, but also highlight its ability to capture semantic information of various objects in the image, resulting in more detailed descriptions of the image pairs. This demonstrates that joint spatial-channel modeling can significantly improve the accuracy and detail of captions generated for the RSICC task.

\section{Conclusions}

In this paper, we propose a novel transformer-based method SAT-Cap for the RSICC task that utilizes joint spatial-channel modeling. The proposed SCE Module effectively addresses the challenges of modeling spatial and channel dependencies in remote sensing images. By sequentially capturing relationships between image regions and the semantic attributes of objects, the SCE enhances feature representation and enables accurate generation of change captions. The integration of depth-wise convolution improves local information modeling, while shared encoder parameters optimize computational efficiency for bi-temporal image pairs. Unlike traditional multi-stage fusion approaches, this method simplifies feature fusion through concatenation and a residual block, achieving precise descriptions of change regions and object transformations.


Extensive experiments have demonstrated that our method can better recognize small objects, regions of image change, and image backgrounds. Although SAT-Cap achieves promising results, several challenges remain:

\begin{enumerate}
    \item SAT-Cap uses the ResNet model as the backbone for feature extraction, rather than leveraging state-of-the-art transformer-based vision models. Therefore, a potential direction for future work is to incorporate advanced backbone models for feature extraction, which could further enhance the model's performance.
    \item The current RSICC task has a limited number of available datasets, and the model has been validated on only two datasets, which may not fully demonstrate its generalization capability. Therefore, a key direction for future work would be the development of higher-quality RSICC datasets to improve model performance and generalization.
\end{enumerate}


\bibliographystyle{IEEEtran}

\bibliography{Ref}

\end{document}